\documentclass[cameraready]{Interspeech}

\title{Evaluating Large Language Models Abilities for\\Addressee, Turn-change, and Next Speaker Prediction in Meetings}

\author[affiliation={1}, correspondingauthor, orcid=0009-0005-6213-3241]{Ryo}{Fukuda} 
\author[affiliation={1}, orcid=0009-0007-1291-8416]{Takatomo}{Kano} 
\author[affiliation={2}, orcid=0000-0003-0375-496X]{Siddhant}{Arora} 
\author[affiliation={1}, orcid=0000-0002-5175-7834]{Marc}{Delcroix} 
\author[affiliation={1}, orcid=0000-0002-1130-5059]{Naohiro}{Tawara} 
\author[affiliation={1}, orcid=0000-0000-0000-1111]{Atsunori}{Ogawa} %
\author[affiliation={1}, orcid=0000-0003-1987-4368]{Yuya}{Chiba} 
\author[affiliation={1}, orcid=0000-0002-3971-0654]{Atsushi}{Ando} 
\author[affiliation={2}, orcid=]{William}{Chen} 
\author[affiliation={2}, orcid=0000-0002-5970-8631]{Shinji}{Watanabe} 

\address{
    $^1$ NTT, Inc., Japan, 
    $^2$ Language Technologies Institute, Carnegie Mellon University, USA
}

\email{ryo.fukuda@ntt.com}

\newcommand{\blue}[1]{\textcolor{blue}{#1}}
\newcommand{\red}[1]{\textcolor{red}{#1}}

\usepackage{comment}
\usepackage{dsfont}
\usepackage[whole]{bxcjkjatype}
\usepackage{multirow}
\usepackage{arydshln}
\usepackage{pifont}
\usepackage{array}

\usepackage{subcaption}

\usepackage{xcolor}
\usepackage{soul}
\usepackage{cite}
\definecolor{foacolor}{RGB}{220,235,255}      
\definecolor{topiccolor}{RGB}{220,255,235}    
\definecolor{rolecolor}{RGB}{240,220,255}     
\newcommand{\hlfoa}[1]{\sethlcolor{foacolor}\hl{\textbf{#1}}}
\newcommand{\hltopic}[1]{\sethlcolor{topiccolor}\hl{\textit{#1}}}
\newcommand{\hlrole}[1]{\sethlcolor{rolecolor}\hl{\textbf{\textit{#1}}}}

\usepackage{url}
\usepackage{cleveref}
\crefname{section}{Section}{Section}
\crefname{figure}{Figure}{Figure}
\crefname{subfigure}{Figure}{Figure}
\crefname{table}{Table}{Table}
\crefname{subtable}{Table}{Table}
\crefname{appendix}{Appendix}{Appendix}

\usepackage{bm}

\usepackage{setspace}

\begin{document}

\maketitle

\begin{abstract}
We investigate turn-taking in multimodal multi-party conversations using large language models (LLMs). We construct an evaluation framework for three tasks: addressee detection, turn-change prediction, and next speaker prediction. We compare supervised models trained for these tasks, text-based LLMs, multimodal LLMs (MM-LLMs), and human subjects. Experiments on the AMI corpus showed that LLMs outperformed supervised models and humans in next speaker prediction, despite not being trained on the target domain and without access to audio or visual information. An MM-LLM performed better than text-based LLMs on addressee detection and turn-change prediction but remained below human performance, indicating difficulty leveraging raw audio-visual signals. Ablation analyses revealed that conversational context was critical, particularly for next speaker prediction. We observed that human and LLM prediction patterns were similar, and intervals with frequent turn changes were difficult for both.
\end{abstract}

\section{Introduction}
Advances in large language models (LLMs) have substantially improved the ability of conversational agents to understand and generate natural language. 
With the emergence of multimodal LLMs (MM-LLMs) capable of processing audio and visual inputs in addition to text~\cite{zhang-etal-2023-video,10448504}, it is becoming possible to integrate linguistic and non-linguistic information to understand non-verbal communication. 
These developments have increased interest in systems that can engage in human multi-party conversations (MPCs), such as meeting assistants and collaborative agents~\cite{matsuyama2013four,skantze2015exploring}, as studied in several MPC corpora~\cite{1198793,Koutsombogera2018-ne,Watanabe2020-bm,Vinnikov2024-an,Nguyen2026-jb}.

Conversational agents participating in MPCs are required to understand turn-taking behaviors, which are more complex than those in dyadic interactions~\cite{traum2003issues}.
In dyadic conversations, understanding turn-taking only requires detecting speaker changes, as the addressee of the current utterance and the next speaker are self-evident (i.e., the interlocutor).
In contrast, MPCs involve multiple potential addressees and multiple candidates for the next speaker.
An utterance may be directed to a specific individual, several participants, or the group as a whole.
Successful participation therefore requires inferring who is being addressed and whether and to whom the conversational floor will shift. 
Failure to appropriately manage turn-taking can result in long pauses or inappropriate interruptions that significantly disrupt the fluidity of the interaction~\cite{johansson2015opportunities}.

Previous studies have extensively investigated turn-taking in dyadic conversations \cite{Skantze2017-ab,roddy2018multimodal,Ekstedt2022-up}, recently examining the turn-taking capabilities of LLMs and audio foundation models~\cite{Ekstedt2020-ds,aroratalking,lin2025full}.
In comparison, turn-taking in MPCs has received less attention, although interest has grown in recent years~\cite{castillo-lopez-etal-2025-survey}.
In MPCs, turn-taking has been studied through individual tasks such as addressee detection~\cite{Jovanovic2004-tw}, turn-change prediction~\cite{laskowski-2010-modeling}, and next speaker prediction~\cite{Ishii2013-me}.
These tasks have been shown to benefit from multimodal information.
For example, visual cues such as gaze contribute to addressee detection and next speaker prediction~\cite{Jovanovic2004-tw,Ishii2013-me}, while prosodic features provide cues for turn-change prediction~\cite{Ohsuga2005-jc}.
Earlier work primarily employed supervised models such as conditional random fields~\cite{10.1145/1647314.1647332}, support vector machines (SVM)~\cite{Malik2019-pb}, and deep neural networks~\cite{Heo2025-oo,Elmers2025-yy}, typically combining textual features with audio-visual signals.
More recently, LLM-based approaches have been explored.
Several studies have evaluated the ability of text-based LLMs to perform addressee detection and next speaker prediction~\cite{Koji2025-zu,Hilgert2025-rl}.
The effectiveness of incorporating visual information, such as gaze or images, to text-based and MM-LLMs has also been investigated~\cite{Koji2025-zu,Mori2026-zq}.

Despite these efforts, three limitations remain.
First, it is unclear to what extent MM-LLMs can perform these tasks directly from raw audio and video signals.
Although such models accept multimodal inputs, their ability to exploit these signals effectively for turn-taking prediction in MPCs has not been systematically examined.
Second, comprehensive evaluations across modeling paradigms, such as supervised models trained to predict these tasks, text-based LLMs, and MM-LLMs, under a unified experimental protocol are limited. 
Third, and most importantly, human performance on these tasks has not been sufficiently quantified under comparable multimodal conditions.
Understanding the inherent difficulty of these tasks for humans and the factors that contribute to it is essential for developing conversational agents that approximate human behavior.

\begin{table*}[t]
\caption{LLM-based approaches for turn-taking in multi-party conversation.}
\vspace{-3mm}
\centering
\footnotesize
\setlength{\tabcolsep}{5pt}
\begin{tabular}{llllll}
\toprule
Study & Data & \# Spk. & Turn-taking tasks & Input modality & Compared approaches \\
\midrule
Inoue et al.~\cite{Koji2025-zu} & TEIDAN~\cite{Koji2025-zu} & 3 & Addr., Next & Text & MM-LLM \\
Hilgert \& Niehues~\cite{Hilgert2025-rl} & AMI, DiPCo~\cite{Van-Segbroeck2020-bj}, ML~\cite{Wei2023-bc} & 3--4 & Next & Text & LLM, \textit{Human} \\
Mori et al.~\cite{Mori2026-zq} & TEIDAN & 3 & Next & Text, Image & MM-LLM \\
Chang et al.~\cite{chang2025multimodal} & TV-MMPC~\cite{chang2025multimodal} & 4.18 (Avg.) & Addr. (Offline) & Text, Audio, Video, Image & MM-LLM \\ \midrule
Ours & AMI (+Multisimo) & 4 & Addr., Turn, Next & Text, Audio, Video & LLM, MM-LLM, \textit{Human} \\
\bottomrule
\vspace{-5mm}
\end{tabular}
\label{tab:related}
\end{table*}

In this study, we address these limitations by evaluating addressee detection, turn-change prediction, and next speaker prediction in a unified framework.
Both models and humans address each task using only past and current conversation information, which reflects real-time conversational constraints. 
We use the AMI Meeting Corpus~\cite{kraaij2005ami}, which consists of natural MPCs in meetings involving four English speakers. 
The corpus provides synchronized audio, video, and manual transcriptions, as well as annotations such as addressee and focus-of-attention (FOA) labels, making it suitable for our experiments.
We compare three classes of models: supervised models, text-based LLMs, and MM-LLMs.
As supervised baselines, we implement four conventional models. For text-based LLMs, we evaluate three variants of Qwen3~\cite{Yang2025-vg}.
For MM-LLMs, we examine Qwen-Omni models~\cite{Xu2025-mb,Xu2025-vy} and Gemini 2.5 Pro.
In addition, we conduct human evaluations under comparable multimodal settings to directly compare humans and models.

The evaluation revealed several key findings.
\begin{itemize}[leftmargin=*]
\item First, human performance on these tasks was not particularly high.\footnote{We should note that a limitation of the study is that, as explained later, our human evaluation was carried out by non-native speakers, who may perform worse than native speakers on these tasks.}
For example, in next speaker prediction, where four candidates were possible, the F1 score was approximately 60\%.
This indicates that turn-taking in MPCs is difficult to predict, even for humans.
\item Second, text-based LLMs outperformed human subjects and supervised models in next speaker prediction.
We confirmed that conversational context was critical for this task, consistent with prior findings~\cite{Hilgert2025-rl}.
\item Third, among MM-LLMs, Gemini 2.5 Pro achieved higher performance than text-based LLMs in addressee detection and turn-change prediction, but still underperformed human performance.
Our analysis indicates that current MM-LLMs still face difficulties in effectively leveraging raw audio-visual signals for turn-taking prediction.
\item Finally, we observed that the prediction tendencies of humans and LLMs were broadly similar.
Intervals that were difficult for humans also tended to be difficult for LLMs.
In particular, segments with frequent speaker changes and more balanced participation among speakers were harder to predict.
\end{itemize}

\section{Related Work}
\subsection{Evaluation of LLMs}
Several recent studies have examined the ability of LLMs to understand turn-taking in MPCs (\cref{tab:related}).
Inoue et al.~\cite{Koji2025-zu} constructed a benchmark for addressee detection and next speaker prediction using three-party conversations.
They reported that LLM performance with ground-truth transcriptions was close to chance level and that incorporating gaze information did not substantially improve performance.
Hilgert and Niehues~\cite{Hilgert2025-rl} evaluated text-based LLMs on next speaker prediction using several MPC datasets.
Their results showed that some models outperformed human subjects and highlighted the importance of conversational context for predicting the next speaker.
Mori et al.~\cite{Mori2026-zq} investigated next speaker prediction using an MM-LLM with text and image inputs on the TEIDAN dataset.
They found no clear performance improvement from incorporating visual information, suggesting that current models may have a limited ability to exploit visual cues for this task.
Chang et al.~\cite{chang2025multimodal} explored whether MM-LLMs can understand conversational structure from audio and video signals.
Using a television dialogue dataset, they evaluated models on discourse-related attributes such as addressee prediction.
However, their evaluation assumes an offline setting in which the entire conversation is available, which differs from online prediction scenarios.

While these studies demonstrate increasing interest in LLM-based approaches to MPC understanding, most focus on a single turn-taking task or rely on limited modalities.
Moreover, systematic comparisons across different model types remain limited.
In contrast, our study focuses on an online scenario simulating an agent participating in a meeting, predicting conversational dynamics without access to future information. Moreover, we compare text-based and MM-LLMs across multiple turn-taking tasks.

\subsection{Human Performance}
Several studies have explored human performance on related turn-taking tasks.
The AMI corpus includes addressee annotations that were carefully produced by trained annotators with access to the entire conversation and guided by dialogue act information.
Inter-annotator agreement for these annotations has been reported to be moderate (Krippendorff's $\alpha$ $\approx$ 0.45–0.56)~\cite{Akker2009-zf}, indicating that addressee identification is inherently ambiguous.
However, in natural conversation, participants infer addressees intuitively, without explicitly referring to dialogue act definitions.
The extent to which such online human judgments align with annotations remains unclear.

Human ability to predict turn changes has also been investigated.
De Ruiter et al.~\cite{de2006projecting} reported that listeners predict turn ends approximately 200 milliseconds before the current speaker finishes.
Casillas and Frank~\cite{casillas2013development} found that even 1–2-year-old children can predict turn changes by integrating lexical and prosodic cues.
However, their experiments were limited to dyadic conversations, and did not quantify utterance-level prediction accuracy.
Hilgert and Niehues~\cite{Hilgert2025-rl} evaluated human performance of next speaker prediction in MPCs, but their evaluation relied solely on textual information. 
This setup may underestimate human performance in natural conversations where multimodal cues such as gaze and prosody are available.

To the best of our knowledge, no prior study has quantitatively evaluated human performance on addressee detection, turn-change prediction, and next speaker prediction under a multimodal and online setting in MPCs.
Our study provides the first unified comparison of humans and models on these tasks, enabling direct assessment of their relative performance.

\section{Task Definition} \label{sec:task}
In this study, we evaluate turn-taking prediction in MPCs through three tasks: (1) \textbf{addressee detection}, (2) \textbf{turn-change prediction}, and (3) \textbf{next speaker prediction}.
In addition to evaluating models, we also measure human performance on the same tasks to clarify the gap between humans and current models for these tasks.

In our experiments, systems perform only two tasks: (1) addressee detection and (2) next speaker prediction at the utterance level, from which turn-change labels can be directly derived.
This formulation simplifies the prediction process while enabling evaluation of different aspects of turn-taking.

\subsection{Addressee detection} \label{subsec:task:address}
Let a conversation be represented as an $N$-length sequence of utterances $(u_1, u_2, \dots, u_N)$.
Each utterance $u_i$ is associated with a speaker ID $s_i \in \mathcal{P}$, a transcription $t_i$, an utterance-level audio segment $a_i$, and an utterance-level video segment $v_i$, where $\mathcal{P}$ denotes the set of participants and $|\mathcal{P}|=K$.
In our experiments using the AMI corpus, \(K=4\).
At time step $i$, the goal is to predict the addressee label
$y_i \in \mathcal{P} \cup \{\texttt{Group}, \texttt{None}\}$ for the current utterance $u_i$.
\texttt{Group} indicates multiple participants, and \texttt{None} indicates that the utterance is not directed to any specific participant.
Addressee detection predicts $y_i$ from $(s_i, t_i, a_i, v_i, c_i)$, where $c_i$ is conversational context derived from preceding utterances.
The form of the conversational context $c_i$ depends on the systems (see \cref{subsec:dataset:feat}).

For evaluation, we report classification accuracy ($\mathrm{Acc}$).
To account for class imbalance, we additionally report the macro-averaged F1 score ($\mathrm{F1}_{\mathrm{ma}}$) over all classes.

\subsection{Turn-change and next speaker prediction} \label{subsec:task:nsp}
We consider two closely related tasks: turn-change prediction and next speaker prediction.
Both tasks are defined based on predicting who will speak immediately after each utterance $u_i$.
Models and humans predict a possibly multi-valued set of next speaker candidates \(\hat{S}_{i+1} \subseteq \mathcal{P}\) from the current inputs $(s_i, t_i, a_i, v_i, c_i)$.\footnote{In a pilot study, subjects often found it difficult to commit to a single next speaker candidate. Therefore, both models and humans are allowed to output one or more candidates.}
\\
\textbf{Turn-change prediction:} \label{subsubsec:task:turn}
Turn-change prediction is the task of predicting whether the speaker will change in the next utterance.
We define the ground-truth turn-change label $r_i$ as
\begin{spacing}{0.8}
\[
r_i =
\begin{cases}
\texttt{Hold} & \text{if } s_{i+1}=s_i,\\
\texttt{Shift} & \text{otherwise}.
\end{cases}
\]
The predicted label is derived from \(\hat{S}_{i+1}\) as
\[
\hat{r}_i =
\begin{cases}
\texttt{Hold} & \text{if } \hat{S}_{i+1}=\{s_i\},\\
\texttt{Shift} & \text{otherwise}.
\end{cases}
\]
\end{spacing}
A \texttt{Hold} is predicted only when the predicted set contains exclusively the current speaker.
For turn-change prediction, we report $\mathrm{Acc}$ and $\mathrm{F1}_{\mathrm{ma}}$ over the two classes.
\\
\textbf{Next speaker prediction:} \label{subsubsec:task:next}
Next speaker prediction is evaluated only on turns with a speaker change (\(r_i\;=\texttt{Shift}\)).
We report a strict accuracy that counts a prediction as correct only when the predicted set contains exactly one speaker and it matches the ground-truth next speaker, i.e., \(\hat{S}_{i+1}\;=\{s_{i+1}\}\).
Because this strict criterion does not account for ambiguous situations, we also report $\mathrm{Precision}$, $\mathrm{Recall}$, and $\mathrm{F1}$, computed based on whether the ground-truth next speaker $s_{i+1}$ is included in $\hat{S}_{i+1}$.

\section{Dataset} \label{sec:dataset}
We constructed an evaluation set for the above tasks.
We used the AMI corpus, which consists of 100 hours of meeting recordings, as in previous studies~\cite{Malik2019-pb,Hilgert2025-rl}.
The AMI corpus provides synchronized audio recordings, video streams, and manual transcriptions.
This corpus includes scenario-based meetings where four participants, each playing different roles in a design team, complete a design project over the course of a day.
A subset of the corpus is annotated with dialogue acts, and within a further subset of these sessions, the dialogue act annotations include manually marked addressee information.
Separately, another subset of the corpus is annotated with FOA, which represents the speaker's gaze target.
FOA labels take one of eight values: one of the four participants, \texttt{table}, \texttt{whiteboard}, \texttt{slidescreen}, or \texttt{Unspecified}.
Further details on the annotations are provided on the AMI corpus website\footnote{\url{https://groups.inf.ed.ac.uk/ami/corpus/annotation.shtml}}.
In the original annotation, addressee labels are assigned to spans of consecutive words.
To obtain utterance-level labels suitable for our tasks, we construct addressee labels $y_i \in \mathcal{P} \cup \{\texttt{Group}, \texttt{None}\}$ for each utterance as follows.
For each word in a given utterance, collect all of the original addressee annotations associated with that word.
If no annotation is found, the utterance is labeled as \texttt{None}.
If one annotation is found, the corresponding addressee label is assigned to that utterance.
If multiple annotations are found, the label \texttt{Group} is assigned.
For turn-change and next speaker prediction in \cref{subsec:task:nsp}, we simply use the speaker label $s_{i+1}$ as the ground truth target for given the utterance $u_i$, in line with previous work~\cite{Hilgert2025-rl}.
Utterances at the end of a session, for which no subsequent speaker exists, are excluded from these evaluations.

For audio, we use close-talking headset microphone recordings for each participant.
Due to speech overlap, the microphone of one speaker occasionally captures the voice of another participant.
In some cases, this leakage makes the identity of the next speaker partially observable.
Manual inspection confirmed such leakage in approximately 8.5\% of utterances.
Because our goal is to evaluate continuous turn-taking prediction in realistic multi-party interactions, we did not exclude these instances.
Only short utterances completely overlapping other utterances (typically fillers or backchannels) were excluded.

\subsection{Data Selection} \label{subsec:dataset:data}

\begin{table}[t]
\caption{Statistics of the dataset. \dag Single refers to labels \textit{A}, \textit{B}, \textit{C} or \textit{D}.}
\footnotesize
\vspace{-3mm}
\centering
\setlength{\tabcolsep}{6pt}
\begin{tabular}{lcc}
\toprule
 & \textbf{Full Set} & \textbf{Human-Eval. Subset} \\
\midrule
\# Sessions & 10 & 2 \\
\# Unique speakers & 16 & 8 \\
\# Utterances & 4051 & 347 \\
Duration (minutes) & 262 & 29 \\
\midrule
\multicolumn{1}{l}{Addressee ratio [\%]} \\
Single\dag/Group/None & 30.7/47.6/21.7 & 49.6/28.2/22.2 \\
\midrule
\multicolumn{1}{l}{Turn-change ratio [\%]} \\
Shift/Hold & 63.7/36.3 & 63.5/36.5 \\
\bottomrule
\end{tabular}
\label{tab:dataset_statistics}
\vspace{-5mm}
\end{table}

We manually inspected the corpus and selected sessions according to the following criteria: (i) availability of addresse and FOA labels, (ii) the camera view clearly captures the faces of all participants, and (iii) availability of headset microphone recordings for all speakers.
As a result, the following 11 sessions remained:
IS1000a, IS1001a, IS1001b, IS1001c, IS1003d, IS1006b, IS1006d, IS1008a, IS1008b, IS1008c, IS1008d.
Of these, ten sessions, excluding IS1000a, which was used for debugging, were used for evaluation.

In addition, we constructed a subset for human evaluation using IS1001a and IS1008a, due to time and resource constraints.
These sessions involve different groups of participants, ensuring diversity in speaker characteristics.
In the AMI corpus, the session suffixes (a, b, c, d) correspond to four phases of the same meeting conducted by a single group (e.g., IS1001 or IS1008).
We therefore selected phase-a sessions, which are the least influenced by preceding conversational context.
\cref{tab:dataset_statistics} summarizes the statistics of the selected data.
Except for a higher proportion of individually addressed interactions, the turn-change ratio and addressee distribution are comparable to those of the full sessions.

\begin{table}[t]
\caption{Summary of input features.}
\vspace{-3mm}
\footnotesize
\centering
\begin{tabular}{ll}
\toprule
Symbol & Description \\
\midrule
$s_i$ & Speaker ID of utterance $u_i$ (ground-truth) \\
$t_i$ & Ground-truth transcription of $u_i$ \\
$a_i$ & Utterance-level audio segment aligned to $u_i$ \\
$v_i$ & Utterance-level video clip aligned to $u_i$ \\
$c_i$ & Conversational context available at step $i$ \\ \midrule
$g_i$ & FOA label aligned to $u_i$ (annotated from video) \\
$\phi_i$ & Text-derived features (e.g., pronouns, length) \\
$\tilde{t}_i$ & ASR transcription derived from audio $a_i$ \\ \bottomrule
\end{tabular}
\label{tab:feature_summary}
\vspace{-6mm}
\end{table}

\subsection{Features} \label{subsec:dataset:feat}
\cref{tab:feature_summary} summarizes the input features. 
In the primary setting, ground-truth $s_i$ and $t_i$, and the utterance-aligned raw signals $a_i$ and $v_i$, are provided.
Speaker IDs are overlaid on the video to associate visual streams with $\mathcal{P}$ (see \cref{fig:mpc:tool}).

The form of the conversation context $c_i$ varies depending on the systems.
For LLMs (\cref{subsec:model:textllm,subsec:model:mmllm}), it consists of speaker-transcription pairs from preceding utterances, $(s_{< i}, t_{< i})$.
For supervised models (\cref{subsec:model:supervised}), following prior work~\cite{Malik2019-pb}, we provide a speaker ID of an preceding utterance, $c_i=s_{i-1}$.
Human subjects (\cref{sec:human-eval}) are not explicitly given contextual information.
However, because they solve the tasks sequentially, they may implicitly rely on contextual information retained in memory from previous utterances.

We additionally consider optional inputs.
For supervised models, the transcription is represented via a text-derived feature vector \(\phi_i\) computed from \(t_i\), rather than raw text.
Specifically, \(\phi_i\) consists of second-person pronoun usage (\textit{you}), presence of conjunctions, and sentence length~(\cite{Malik2019-pb} for details).
To assess robustness under practical conditions, we also consider an ASR setting in which \(t_i\) is replaced with the ASR transcript \(\tilde{t}_i\) (see \cref{subsec:res:asr}).
We also evaluated a modular setting in which FOA labels \(g_i\) are provided as visual cues~(\cref{subsec:res:foa}).

\begin{table}[t]
\centering
\footnotesize
\setlength{\tabcolsep}{4pt}
\caption{Prompt used for the MM-LLM (Next Speaker Prediction). The prompt was slightly edited for brevity.}
\vspace{-3mm}
\begin{tabular}{p{0.97\linewidth}}
\toprule
\textbf{Description}: You are an expert model of multi-party conversation analysis. Your task is to predict who will speak next in a meeting with four participants: A, B, C, and D.
You are given the current speaker ID, the current utterance (transcription), and an audio-visual video clip, optionally with conversation context (previous speakers and utterances) and focus of attention (where the current speaker is visually attending).
The video shows four participants in a meeting room. Seating positions: A (back right), B (back left), C (front right), D (front left). Video frames are overlaid with participant labels (A–D). Participants may stand or move; use overlaid labels rather than absolute positions.
You may use visual cues (gaze direction, head orientation, posture, gestures) and prosodic cues (intonation, pauses, turn-final cues) to infer the next speaker.
The next speaker may be the same as the current speaker. You may predict one or multiple candidates if ambiguous. \\
\textbf{Output}: One or more labels from \{A, B, C, D\}. \\
\textbf{Context (Speaker: Transcription)}: \\
~~C: Oops. \\
~~C: That's as far as it goes. \\
~~A: Hi guys uh good morning everybody here. \\
\textbf{Current speaker}: A. \\
\textbf{Transcription}: I am a project manager for this new project which we are going to discuss now. \\
\textbf{Focus of attention}: Table. \\

\textbf{Video}: (video clip path) \\
\bottomrule
\end{tabular}
\label{tab:prompt}
\vspace{-6mm}
\end{table}

\section{Model Evaluation} \label{sec:model-eval}
We evaluate three classes of models: conventional supervised learning models, and off-the-shelf text-based and MM-LLMs.
As a naive baseline, we report majority or chance-level strategies for each task.
For addressee detection, the naive baseline always predicts \texttt{Group} label.
For turn-change prediction, it always predicts \texttt{Shift} label.
For next speaker prediction, we report the expected accuracy of random selection among the four participants, which is 25.0.

\subsection{Supervised Learning Models} \label{subsec:model:supervised}
We implemented four supervised models used in prior work of addressee detection~\cite{Malik2019-pb}:
Naive Bayes~\cite{rish2001empirical}, Random Forest~\cite{liaw2002classification}, Multi-layer Perceptron (MLP)~\cite{kruse2022multi}, and Support Vector Machine (SVM)~\cite{steinwart2008support}.
Hyperparameters were set following~\cite{Malik2019-pb}.
The input consists of the current speaker $s_i$, the text-derived features $\phi_i$, and the context $c_i=s_{i-1}$.

All models were implemented using scikit-learn~\cite{kramer2016scikit}.
We conducted five-fold cross-validation over the ten sessions using \texttt{GroupKFold}, ensuring that utterances from the same session did not appear in both training and test sets.
Predictions on the test split of each fold were aggregated, and performance metrics were computed over the combined test data.

\subsection{LLMs} \label{subsec:model:textllm}
As LLMs, we evaluated three variants of Qwen3~\cite{Yang2025-vg}, a recently proposed model with the Transformer-based decoder architecture~\cite{vaswani2017attention}: Qwen3-8B, Qwen3-14B, and Qwen3-32B.
For these models, the thinking mode was enabled by default, and its contribution was examined through an ablation study.
Inference was performed in bfloat16 precision using greedy decoding.
For each utterance $u_i$, text-based LLMs were given the task instruction and the text-based inputs $(s_i, t_i, c_i)$.
The prompt for text-based LLMs is the same as that for MM-LLMs described below, except for the absence of audio-visual inputs $(a_i, v_i)$.

\subsection{MM-LLMs} \label{subsec:model:mmllm}
We evaluate MM-LLMs that jointly process text, audio, and video inputs.
For each utterance $u_i$, the models were given the task instruction and the multimodal inputs $(s_i, t_i, a_i, v_i, c_i)$.
\cref{tab:prompt} presents an example prompt for next speaker prediction.

We evaluate recent open-weight end-to-end multimodal foundation models, Qwen2.5-Omni~\cite{Xu2025-mb} and Qwen3-Omni~\cite{Xu2025-vy}, which have demonstrated strong performance in both unimodal and multimodal understanding tasks.
These models adopt a Thinker–Talker architecture, where the Thinker module is responsible for text generation and reasoning, and the Talker module generates streaming speech tokens.
In this study, only the Thinker module was used.
From the Qwen2.5-Omni family, we selected Qwen2.5-Omni-7B, a 7B-parameter model.
For Qwen3-Omni, we evaluated Qwen3-Omni-30B-A3B-Thinking (Qwen3-Omni-30B), a 30B-parameter mixture-of-experts model optimized for reasoning.
The instruction-tuned variant (Qwen3-Omni-30B-A3B-Instruct) was excluded because preliminary experiments showed inferior performance compared to the Thinking model.
Inference was performed in bfloat16 precision using greedy decoding.

We also evaluated the closed-source multimodal model Gemini 2.5 Pro via the official API.
The same task instructions and input features were provided to the local multimodal models.
Inference was conducted with a temperature set to 1.0.

\begin{figure}[t]
    \centering
    \includegraphics[width=0.85\linewidth]{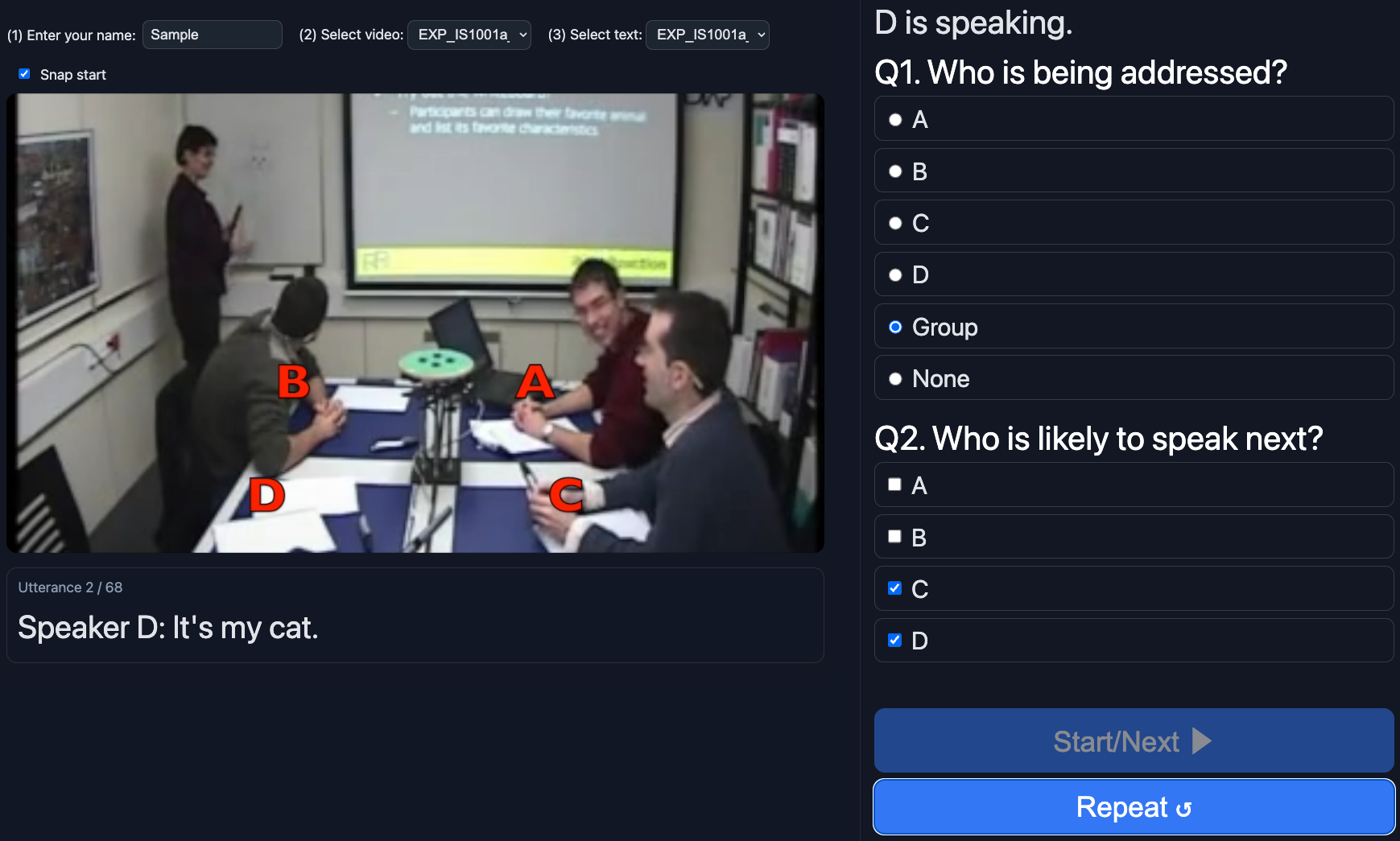}
    \vspace{-2mm}
    \caption{Screenshot of the experiment tool.}
    \label{fig:mpc:tool}
    \vspace{-5mm}
\end{figure}

\section{Human Evaluation} \label{sec:human-eval}
To compare human and model performance, we conducted a human evaluation under the same task formulation described in \cref{sec:task}.
Participants simultaneously performed addressee detection, turn-change prediction, and next speaker prediction in an online setting, without access to future utterances.

We developed a web-based interface (\cref{fig:mpc:tool}) that plays videos and audio from the beginning and automatically pauses them at the end of each utterance.
At each pause for utterance $u_i$, subjects were given the current speaker ID and the transcription, together with the corresponding audio and video clips, i.e., $(s_i, t_i, a_i, v_i)$.
They then answered two questions: \textit{Q1. Who is being addressed?} and \textit{Q2. Who is likely to speak next?}
Formally, they predicted the addressee label $y_i$~(\cref{subsec:task:address}) and a set of next speaker candidates $\hat{S}_{i+1} \subseteq \mathcal{P}$~(\cref{subsec:task:nsp}).
After answering, playback resumed until the next utterance boundary.

Subjects were instructed to respond based on their intuition.
To ensure consistency in the annotation criteria, all participants engaged in a practice session before the main experiment and were provided with example responses.
To reduce participant burden, each session was divided into segments of approximately five minutes.
Thus, participants could rely on at most about five minutes of prior conversational context when making predictions.
For a fair comparison, when evaluating models on the human-evaluation subset, we used the same segmented session chunks, rather than the full sessions.
Twelve non-native English speakers who regularly use English in professional settings participated in the experiment.
Each utterance received three independent annotations.
It should be noted that our experimental design differs from natural conversational settings in several aspects and may underestimate human performance.
We discuss these limitations in \cref{sec:limitation}.

Inter-participant agreement, measured by Krippendorff's $\alpha$, was 0.67 for addressee detection and 0.48 for next speaker prediction.
The agreement for addressee detection indicates a moderate level of consistency among participants, whereas the lower agreement for next speaker prediction reflects the inherent ambiguity of anticipating future turns in MPCs.
The agreement for addressee detection is slightly higher than the previously reported inter-annotator agreement for AMI addressee labels (approximately 0.45–0.56)~\cite{Akker2009-zf}.
This may seem counterintuitive, given that the original annotations were produced by trained annotators under stricter guidelines.
One possible explanation is the difference in labeling granularity: AMI annotations were assigned at the word level, whereas our evaluation was conducted at the coarser utterance level, which may have resulted in higher agreement.
For ext speaker prediction, our agreement is substantially higher than the Fleiss' kappa of 0.17 reported in prior work~\cite{Hilgert2025-rl}.
This difference may stem from experimental settings, as the previous study relied only on textual information, whereas our participants had access to multimodal cues.

\begin{table}
    \setlength{\tabcolsep}{3pt}
    \footnotesize
    \caption{Overall results on the full set. Input features: $a_i$ (audio segment), $v_i$ (video clip), $s_i$ (speaker ID), $t_i$ (transcription), and $c_i=(s_{< i}, t _{< i})$ (conversational context). $\mathrm{Acc}$ was reported for Addressee detection (Addr.) and Turn-change prediction (Turn), and $\mathrm{F1}$ for Next speaker prediction (Next). The naive baseline corresponds to predicting the majority class for addressee detection (\texttt{Group}) and turn-change prediction (\texttt{Shift}), and random selection for next speaker prediction.}
    \vspace{-3mm}
    \centering
    \begin{tabular}{lccccccccc} \toprule
         & \multicolumn{2}{c}{\textbf{Raw}} & \multicolumn{3}{c}{\textbf{Annotated}} 
         & \multirow{2}{*}{\begin{tabular}[c]{@{}c@{}}\textbf{Addr.}\\($\mathrm{Acc}$)\end{tabular}}
         & \multirow{2}{*}{\begin{tabular}[c]{@{}c@{}}\textbf{Turn}\\($\mathrm{Acc}$)\end{tabular}}
         & \multirow{2}{*}{\begin{tabular}[c]{@{}c@{}}\textbf{Next}\\($\mathrm{F1}$)\end{tabular}}\\
        \cmidrule(l{\tabcolsep}r{\tabcolsep}){2-3}
        \cmidrule(l{\tabcolsep}r{\tabcolsep}){4-6}
         & $a_i$ & $v_i$ & $s_i$ & $t_i$ & $c_i$  \\ \midrule
        \textit{Naive baseline} &  &  &  &  &  & 47.6 & 63.7 & 25.0 \\ \midrule
        \textit{Supervised model} \\
        ~~Naive Bayes &  &  & \checkmark & \checkmark & \checkmark & 47.6 & 66.6 & 40.3 \\
        ~~Random Forest &  &  & \checkmark & \checkmark & \checkmark & 49.4 & 60.2 & 31.0 \\
        ~~MLP &  &  & \checkmark & \checkmark & \checkmark & 55.3 & 66.3 & 40.0 \\
        ~~SVM &  &  & \checkmark & \checkmark & \checkmark & \textbf{56.4} & 66.5 & 40.1 \\ \midrule
        \textit{LLM} \\
        ~~Qwen3-8B &  &  & \checkmark & \checkmark & \checkmark & 51.8 & 64.6 & 50.0 \\
        ~~Qwen3-14B &  &  & \checkmark & \checkmark & \checkmark & 52.3 & 66.4 & \textbf{51.1} \\
        ~~Qwen3-32B &  &  & \checkmark & \checkmark & \checkmark & 47.3 & 65.3 & 49.7  \\ \midrule
        \textit{MM-LLM} \\
        ~~Qwen2.5-Omni-7B & \checkmark & \checkmark & \checkmark & \checkmark & \checkmark & 50.0 & 61.6 & 37.0 \\
        ~~Qwen3-Omni-30B & \checkmark & \checkmark & \checkmark & \checkmark & \checkmark & 32.1 & 41.6 & 13.7 \\
        ~~Gemini 2.5 Pro & \checkmark & \checkmark & \checkmark & \checkmark & \checkmark & 55.7 &  \textbf{68.3} & 47.7 \\ \bottomrule
    \end{tabular}
    \label{tab:main}
    \vspace{-2mm}
\end{table}

\begin{table}
    \caption{Comparison of models and humans in the subset. \dag Significantly better than \textit{Human}; \ddag significantly worse than \textit{Human}~(paired bootstrap, 95\% CI).}
    \vspace{-3mm}
    \setlength{\tabcolsep}{3pt}
    \centering
    \resizebox{\linewidth}{!}{
    \begin{tabular}{lcccccccc} \toprule
         & \multicolumn{2}{c}{\textbf{Addressee }} & \multicolumn{2}{c}{\textbf{Turn-change }} & \multicolumn{4}{c}{\textbf{Next speaker }} \\
        &  \multicolumn{2}{c}{\textbf{ detection}} & \multicolumn{2}{c}{\textbf{prediction}} & \multicolumn{4}{c}{\textbf{ prediction}} \\
        \cmidrule(l{\tabcolsep}r{\tabcolsep}){2-3}\cmidrule(l{\tabcolsep}r{\tabcolsep}){4-5}\cmidrule(l{\tabcolsep}r{\tabcolsep}){6-9}
         & $\mathrm{Acc}$ &  $\mathrm{F1}_{\mathrm{ma}}$ & $\mathrm{Acc}$ & $\mathrm{F1}_{\mathrm{ma}}$ & $\mathrm{Acc}$ & $\mathrm{P}$ & $\mathrm{R}$ & $\mathrm{F1}$ \\ \midrule
        \textit{Naive baseline} & 28.4 & 7.4 & 64.6 & 39.3 & 25.0 & 25.0 & 25.0 & 25.0 \\ \midrule
        \textit{Human} & \textbf{66.6} & \textbf{67.5} & \textbf{75.0} & \textbf{71.5} & 47.2 & 52.2 & 70.7 & 60.1 \\ \midrule
        Qwen3-14B & 51.5\ddag & 52.0\ddag & 67.4 & 53.8\ddag & \textbf{53.7} & \textbf{60.3} & \textbf{81.8} & \textbf{69.4}\dag \\
        Gemini 2.5 Pro & 61.4 & 57.4 & 69.8 & 60.6\ddag & 45.3 & 50.7 & 66.8 & 57.7 \\ \bottomrule
    \end{tabular}}
    \label{tab:human_model}
    \vspace{-5mm}
\end{table}

\section{Results} \label{sec:results}
\subsection{Model comparison} \label{subsec:res:main}
\textbf{Supervised models vs. LLMs:}
\cref{tab:main} shows the overall performance of models.
SVM achieved the highest accuracy in addressee detection.
In turn-change prediction, it also outperformed all LLMs except Gemini 2.5 Pro.
These results indicate that task-specific supervised models can surpass general LLMs in these tasks, even without multimodal information.
In contrast, all text-based LLMs and Gemini 2.5 Pro outperformed supervised models in next speaker prediction.
This task depends on the conversational context~\cite{Hilgert2025-rl}.
Unlike supervised models, which rely only on features from the immediately preceding utterance as context, LLMs can leverage longer conversational histories, likely contributing to their advantage in this task.
\\
\textbf{LLM model size:}
Among text-based LLMs, we found no monotonic correlation between model size and performance.
Qwen3-14B outperformed both the smaller 8B and larger 32B variants across tasks.
This suggests that the 14B model has a certain level of reasoning ability to solve these tasks, and that the additional parameters of the 32B model do not necessarily yield further gains.
For example, in addressee detection, the 32B model more frequently predicted individual classes (A–D) when the correct label was the majority class \texttt{Group}, leading to reduced accuracy.
This may reflect the tendency of the 32B model to prefer detailed predictions over simple answers.
\\
\textbf{Comparison with prior work:}
Text-based LLMs outperformed the naive baseline in both addressee detection and next speaker prediction.
This contrasts with the findings of Inoue et al.~\cite{Koji2025-zu}, who reported near-chance performance of GPT-4o on similar tasks in Japanese three-party conversations.
The discrepancy may stem from differences in models, language, and domain (Japanese three-party goal-free discussions vs. English four-party meetings), and the amount of contextual input provided (e.g., five utterances vs. entire conversation history).
Investigating how task difficulty varies across languages and domains remains an important direction for future research.
\\
\textbf{MM-LLM:}
The local MM-LLMs, Qwen2.5-Omni-7B and Qwen3-Omni-30B, did not outperform the SVM or the text-based LLMs.
This suggests that the audio-visual features were not effectively leveraged in these models.
In contrast, Gemini 2.5 Pro obtained the highest scores among LLMs in addressee detection and turn-change prediction.
We discuss the factors underlying this strong performance of Gemini 2.5 Pro in~\cref{sec:analysis}.
In the next speaker prediction, which depends more strongly on conversational context, Gemini 2.5 Pro performed worse than the text-based LLMs.
Possibly, incorporating audio-visual inputs increased the complexity of input, making effective contextual reasoning more difficult, as discussed in~\cref{sec:analysis}.

\begin{table}[t]
    \setlength{\tabcolsep}{3pt}
    \caption{Ablation results for Qwen3-14B. Input features: $a_i$ (audio segment), $v_i$ (video clip), $s_i$ (speaker ID), $t_i$ (transcription), $\tilde{t}_i$ (ASR transcription), $c_i$ (context), and $g_i$ (FOA). $\diamond$: Setting without thinking mode. \dag~Significantly better than (a); \ddag~significantly worse than (a).}
    \footnotesize
    \vspace{-3mm}
    \renewcommand{\arraystretch}{1.1}
    \centering
    \begin{tabular}{llccccccc ccc} \toprule
         & \multirow{2}{*}{\textbf{Setting}} & \multicolumn{2}{c}{\textbf{Raw}} & \multicolumn{5}{c}{\textbf{Annotated}} & \multirow{2}{*}{\begin{tabular}[c]{@{}c@{}}\textbf{Addr.}\\($\mathrm{Acc}$)\end{tabular}} & \multirow{2}{*}{\begin{tabular}[c]{@{}c@{}}\textbf{Turn}\\($\mathrm{Acc}$)\end{tabular}} & \multirow{2}{*}{\begin{tabular}[c]{@{}c@{}}\textbf{Next}\\($\mathrm{F1}$)\end{tabular}}\\
        \cmidrule(l{\tabcolsep}r{\tabcolsep}){3-4}\cmidrule(l{\tabcolsep}r{\tabcolsep}){5-9}
        & & $a_i$ & $v_i$ & $s_i$ & $t_i$ & $\tilde{t}_i$ & $c_i$ & $g_i$ \\ \midrule
        (a) & Primary & & & \checkmark & \checkmark & &\checkmark & & 52.3 & 66.4 & 51.1 \\
        (b) & + ASR & & & \checkmark & & \red{$\hat{\checkmark}$} & \checkmark & & 51.2 & 65.6 & 51.1 \\
        (c) & + FOA & & & \checkmark & \checkmark & &\checkmark & \blue{\checkmark} & 55.0 & 65.8 & 67.7 \\ \hline\hline
        (d) & -- Context & & & \checkmark & \checkmark & & $\red\times$ & & 42.5\ddag & 56.2\ddag & 36.9\ddag \\ \hline\hline
        (e) & -- Think$\diamond$ & & & \checkmark & \checkmark & & \checkmark & & 47.3\ddag & 64.9 & 48.1\ddag \\\bottomrule
    \end{tabular}
    \label{tab:ablation_qwen3}
    \vspace{-2mm}
\end{table}

\begin{table}[t]
    \setlength{\tabcolsep}{3pt}
    \caption{Ablation results for Gemini 2.5 Pro in the subset.} 
    \footnotesize
    \vspace{-3mm}
    \renewcommand{\arraystretch}{1.1}
    \centering
    \begin{tabular}{llccccccc ccc} \toprule
         & \multirow{2}{*}{\textbf{Setting}} & \multicolumn{2}{c}{\textbf{Raw}} & \multicolumn{5}{c}{\textbf{Annotated}} & \multirow{2}{*}{\begin{tabular}[c]{@{}c@{}}\textbf{Addr.}\\($\mathrm{Acc}$)\end{tabular}} & \multirow{2}{*}{\begin{tabular}[c]{@{}c@{}}\textbf{Turn}\\($\mathrm{Acc}$)\end{tabular}} & \multirow{2}{*}{\begin{tabular}[c]{@{}c@{}}\textbf{Next}\\($\mathrm{F1}$)\end{tabular}}\\
        \cmidrule(l{\tabcolsep}r{\tabcolsep}){3-4}\cmidrule(l{\tabcolsep}r{\tabcolsep}){5-9}
        & & $a_i$ & $v_i$ & $s_i$ & $t_i$ & $\tilde{t}_i$ & $c_i$ & $g_i$ \\ \midrule
        (a) & Primary & \checkmark & \checkmark & \checkmark & \checkmark & & \checkmark & & 61.4 & 69.8 & 57.7 \\
        (b) & + ASR & \checkmark & \checkmark & \checkmark & & \red{$\hat{\checkmark}$} & \checkmark & & 59.9 & 71.3 & 60.9 \\
        (c) & + FOA & \checkmark & \checkmark & \checkmark & \checkmark & & \checkmark & \blue{\checkmark} & 64.1 & 71.3 & 62.6 \\ \hline\hline 
        (d) & -- Context & \checkmark & \checkmark & \checkmark & \checkmark & & $\red\times$ & & 38.3\ddag & 67.1 & 25.7\ddag \\
        (e) & -- Text & \checkmark & \checkmark & \checkmark & $\red\times$ & & $\red\times$ & & 38.6\ddag & 66.2 & 31.1\ddag \\
        (f) & -- Audio & $\red\times$ & \checkmark & \checkmark & \checkmark & & \checkmark & & 61.1 & 71.0 & 58.5 \\
        (g) & -- Video & $\red\times$ & $\red\times$ & \checkmark & \checkmark & & \checkmark & & 59.6 & 69.5 & 64.8 \\ \bottomrule
    \end{tabular}
    \label{tab:ablation_gemini}
    \vspace{-5mm}
\end{table}

\subsection{Humans vs. Models} \label{subsec:res:subset}
\cref{tab:human_model} compares human performance with the best-performing LLMs in the subset.
Human performance was computed by aggregating predictions across all participants.
For addressee detection and turn-change prediction, humans significantly outperformed both Qwen3-14B and Gemini 2.5 Pro.
Multimodal cues such as gaze direction and prosody are known to play an important role in these tasks~\cite{Jovanovic2004-tw,Ishii2013-me,Ohsuga2005-jc}, and human subjects may have been able to more effectively utilize these features, resulting in higher accuracy.
In contrast, for next speaker prediction, the text-based LLM achieved higher performance, and the MM-LLM achieved performance comparable to humans.
Hilgert and Niehues~\cite{Hilgert2025-rl} similarly reported that text-based LLMs outperformed humans in next speaker prediction under text-only conditions.
Our results are consistent with their findings and further show that this advantage persists even when humans have access to multimodal information.
This may be attributed to their stronger ability to model conversational context $c_i = (s_{<i}, t_{<i})$, which appears to be critical for identifying the next speaker, as discussed in~\cref{sec:analysis}.
Note that both humans and LLMs were evaluated under the same constraint described in \cref{sec:human-eval}, i.e., the available preceding context was limited to that corresponding to at most about five minutes of video.

Importantly, these results and moderate inter-participant agreement (reported in \cref{sec:dataset}) indicate that all three tasks, particularly next speaker prediction, are challenging even for humans.
To better understand what makes these tasks difficult, we further analyze how prediction performance varies across conversational segments in \cref{subsec:ana:tempo}.

\subsection{Effect of estimated transcriptions} \label{subsec:res:asr}
The primary setting assumed access to ground-truth speaker ID ($s_i$) and manual transcriptions ($t_i$).
To evaluate model robustness under more practical conditions, we replaced manual transcriptions with ASR outputs ($\tilde{t}_i$).
We used  Whisper large-v3~\footnote{\url{https://huggingface.co/openai/whisper-large-v3}}~\cite{radford2022robustspeechrecognitionlargescale} to transcribe the individual headset microphones, simulating a meeting scenario in which each participant uses an individual microphone.
The word error rate was 24.12 \%.
For Qwen3-14B (\cref{tab:ablation_qwen3} (a) vs. (b)), replacing manual transcripts with ASR slightly decreased performance in addressee detection and turn-change prediction, while next speaker prediction performance remained largely unchanged.
For Gemini 2.5 Pro (\cref{tab:ablation_gemini} (a) vs. (b)), addressee detection performance slightly decreased, while turn-change prediction and next speaker prediction improved~\footnote{Due to cost constraints, the ablation study for Gemini 2.5 Pro was conducted on the subset only.}.
Despite the relatively high WER, the overall impact was limited, and in some cases performance improved.
Many ASR errors resulted from the omission of fillers and minor disfluencies that were included in the ground-truth transcripts.
While these omissions contributed to the WER, they often had little impact on the underlying meaning of the utterances.
These results demonstrate to some extent the robustness of the LLM-based approach in practical environments.
However, in more general scenarios without individual headset microphones, i.e., where speaker diarization and multi-talker ASR are required, these tasks may be more challenging.
Evaluating performance under such conditions
is part of our future work.

\subsection{Effect of gaze information} \label{subsec:res:foa}
We also evaluated a modular setting in which FOA labels ($g_i$) were provided as explicit visual cues.
For both Qwen3-14B (\cref{tab:ablation_qwen3} (a) vs. (b)) and Gemini 2.5 Pro (\cref{tab:ablation_gemini} (a) vs. (b)), incorporating FOA consistently improved performance in addressee detection and next speaker prediction.
This finding aligns with prior research showing that gaze information is a useful signal for identifying the addressee and anticipating the next speaker~\cite{Jovanovic2006-xa,Ishii2013-me}.
In contrast, the effect of FOA on turn-change prediction was mixed.
Performance slightly decreased for Qwen3-14B but slightly improved for Gemini 2.5 Pro.
Prior work has shown that turn-change prediction is more strongly associated with gaze transition patterns rather than the speaker’s gaze direction at a given moment.
Since FOA represents only the current gaze target at each utterance, it may not capture these dynamic transition patterns.

Gemini 2.5 Pro showed improved performance when FOA was added, despite already receiving raw audio and video inputs.
These gains suggest that the model does not fully extract or utilize gaze cues from the raw video.
Thus, explicitly annotated visual attention appears to provide complementary information even for an MM-LLM.
We analyze the contribution of multimodal cues in this model in the next sections.

\begin{figure}[t]
    \centering
    \includegraphics[width=0.8\linewidth]{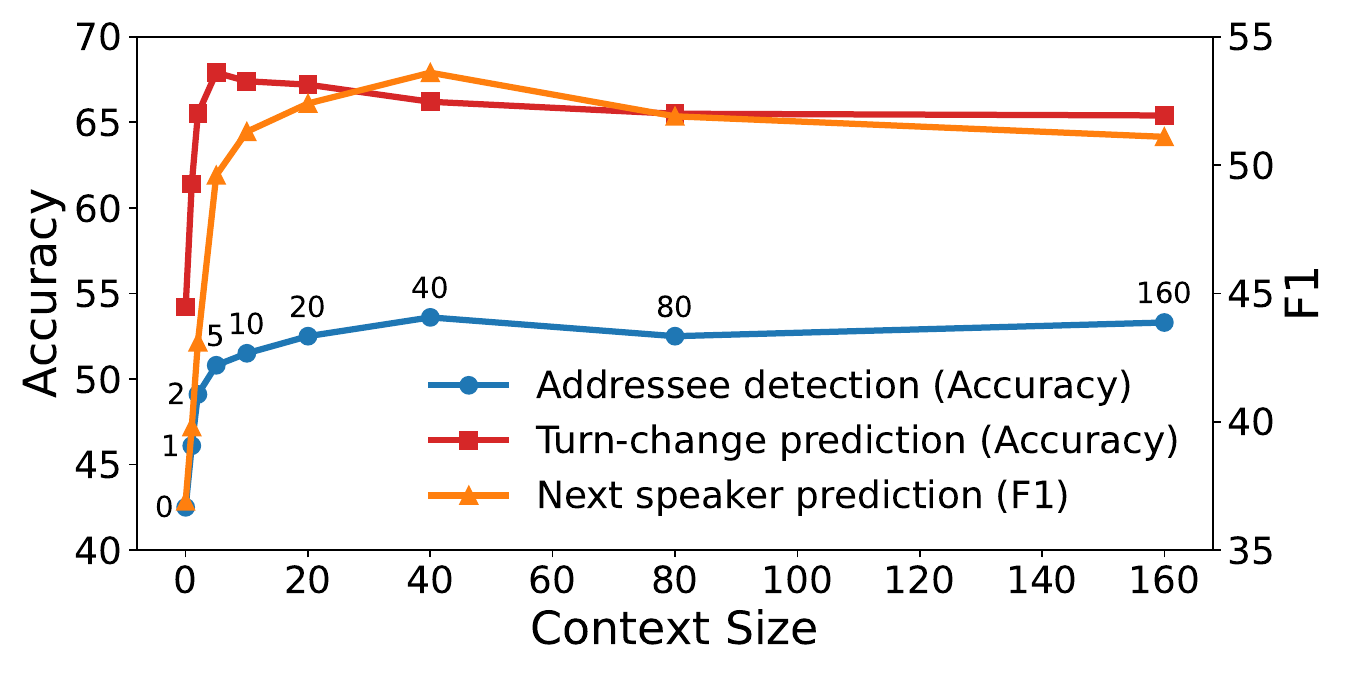}
    \vspace{-4mm}
    \caption{Effect of context size on Qwen3-14B performance.}
    \label{fig:context_effect}
    \vspace{-5mm}
\end{figure}

\section{Analysis} \label{sec:analysis}
\subsection{Important features} \label{subsec:ana:ablation}
\cref{tab:ablation_qwen3,tab:ablation_gemini} include ablation studies examining the contribution of input features.
Firstly, removing conversational context ((a) vs. (d) in \cref{tab:ablation_qwen3,tab:ablation_gemini}) led to a substantial performance degradation for Qwen3-14B and Gemini 2.5 Pro, particularly in addressee detection and next speaker prediction.
These results indicate that conversational context is important in all three tasks.
\cref{fig:context_effect} shows the performance of Qwen3-14B with different context sizes, $C = \{0, 1, 2, 5, 10, 20, 40, 80, 160\}$ utterances.
Introducing even a very short context ($C=1$ or $C=2$) led to large gains over the no-context setting across all tasks.
For addressee detection and next speaker prediction, performance continued to increase as context expanded up to $C=40$, after which it was largely maintained or slightly declined.
In contrast, turn-change prediction peaked at $C=5$ and did not benefit from longer context.
\textbf{These results suggest that addressee detection and next speaker prediction rely more heavily on longer conversational context, whereas turn-change prediction depends primarily on local cues.}
\textbf{Moreover, the improvement with increasing context was most pronounced for next speaker prediction, indicating a particularly strong dependence on contextual information.
}

Secondly, we examined the ablation of transcriptions in the MM-LLM (\cref{tab:ablation_gemini} (e)), where predictions relied only on raw audio-visual signals and the current speaker ID.
Compared to removing context alone (\cref{tab:ablation_gemini} (d)), performance for addressee detection and turn-change prediction was comparable, and next speaker prediction was slightly improved.
This indicates that textual transcription is not the sole source of information for these tasks.
However, it is difficult to determine how much the model relies on raw audio-visual cues, as the speaker ID may already provide a strong prior.
When audio input was removed (\cref{tab:ablation_gemini} (f)), performance changed only marginally across tasks.
Similarly, removing both audio and video (\cref{tab:ablation_gemini} (g)) resulted in comparable or even higher performance for next speaker prediction.
\textbf{These results suggest that, in the current setting, raw audio-visual signals do not consistently provide complementary information beyond text and speaker identity.}
\textbf{Overall, while multimodal inputs are available, their effective integration with textual and speaker information remains limited in the present model.}

\begin{table}[t]
\centering
\footnotesize
\setlength{\tabcolsep}{4pt}
\caption{Example of thought trace of Qwen3-14B with focus-of-attention in the next speaker prediction task.}
\vspace{-3mm}
\begin{tabular}{@{}c@{}}
\toprule
\begin{minipage}[t]{\linewidth}
\setlength{\baselineskip}{0.8\baselineskip}
\setlength{\lineskip}{0pt}
Okay, let's see. The current speaker is D, and their transcription is "Yes yes is it like that." \hlfoa{The focus of attention is the whiteboard.} I need to figure out who is likely to speak next.

\hltopic{Looking at the conversation context, there's been a discussion about drawing animals on the whiteboard. B drew an orangutan, and now D is responding to something, possibly about the drawing. The last few lines involve D saying it's a cat and others reacting.}

Since D just confirmed something about the drawing (maybe agreeing it's a cat), the next speaker might be someone reacting to that. \hlrole{A has been leading the meeting, so maybe A would continue. B was discussing the orangutan earlier, so might comment again. C hasn't spoken much lately. D might not speak again immediately if they just confirmed.} 

\hlfoa{Also, the focus is on the whiteboard, so someone might be pointing out something else on it.} A might want to move the meeting forward, or B could have another comment. C might have an opinion but hasn't been active. 

So possible candidates are A or B. Maybe both. I'll go with A and B.

\end{minipage}
\\
\bottomrule
\end{tabular}
\label{tab:example_qwen3_thinking}
\vspace{-6mm}
\end{table}

\subsection{Reasoning ability of LLMs} \label{subsec:ana:cot}
\cref{tab:example_qwen3_thinking} shows an example of the thought trace of Qwen3-14B with FOA labels for the next speaker prediction task.
This example provides some insights about the predictive capabilities of LLMs.
The model leverages the broader conversational context to infer the flow of the discussion (\hltopic{text in italics}), the roles of participants, and their speaking tendencies  (\hlrole{text in bold italics}), rather than relying only on the most recent utterance.
It also explicitly incorporates the FOA labels to ground its situational reasoning (\hlfoa{text in bold}), suggesting that visual information is actively used in the prediction.
These behaviors may partly explain why LLMs sometimes outperform human subjects in our experiment.
Humans were not given the text-based context, and they took up to twice the duration of the original video to complete the task.
\textbf{In contrast, the LLM had access to the long context for each utterance and could perform step-by-step reasoning, which likely contributed to its higher accuracy.}
As shown in \cref{tab:ablation_qwen3} (e), disabling the thinking mode significantly degraded the performance, supporting this hypothesis.

\subsection{Task difficulty} \label{subsec:ana:tempo}
\begin{figure}[t]
    \centering
    \includegraphics[width=0.95\linewidth]{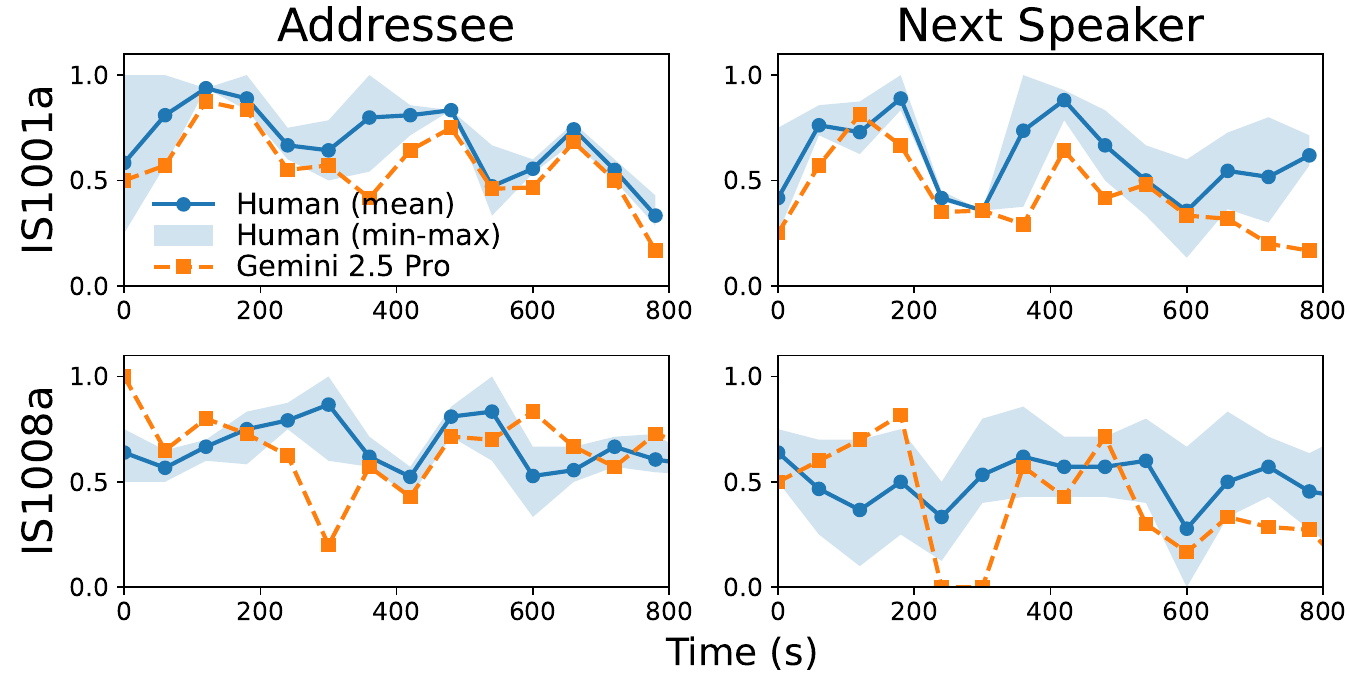}
    \vspace{-3mm}
    \caption{Temporal variation of addressee detection and next speaker prediction accuracies across 1-minute windows. Shaded region indicates the min–max range across participants.}
    \label{fig:temporal_graph}
    \vspace{-2mm}
\end{figure}

In this section, we examine how task difficulty varies over time and how it relates to prediction accuracy.
\cref{fig:temporal_graph} visualizes the temporal changes in addressee detection and next speaker prediction accuracy for humans and Gemini 2.5 Pro using one-minute windows.
\textbf{There are similar trends in the temporal changes of human and Gemini performance.
}
In both sessions, there appear to be ``easy" intervals where accuracy is high for both, and ``difficult" intervals where performance drops.
This indicates that task difficulty fluctuates considerably across conversational intervals within the dataset, and suggests promising directions for future research.
\textbf{Instead of aiming for uniform accuracy across an entire dialogue, we should design models that account for the inherent uncertainty in human conversational dynamics, improving reliability in predictable moments, while allowing for more fluid, stochastic behavior for ambiguous cases.}

\begin{figure}[t]
    \centering
    \includegraphics[width=0.95\linewidth]{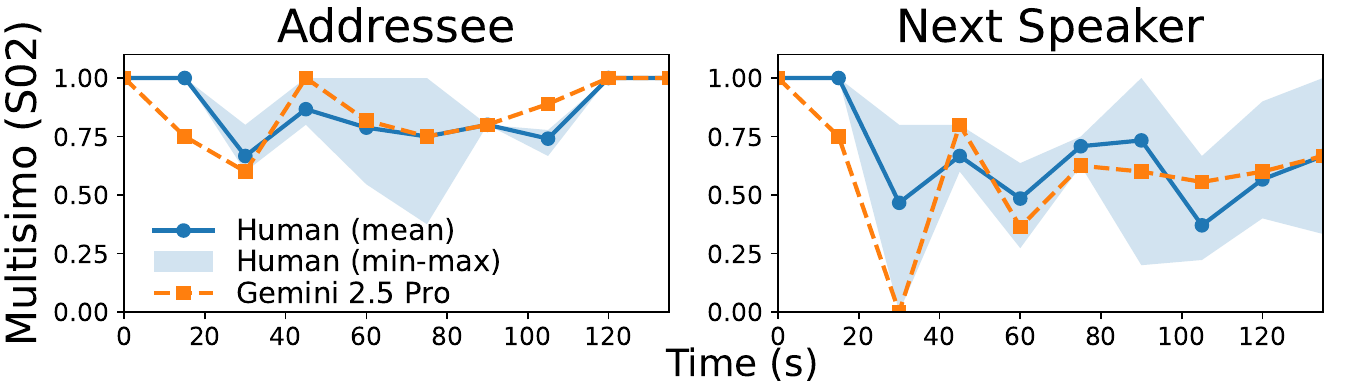}
    \vspace{-3mm}
    \caption{Temporal variation on the Multisimo subset using 15-second windows.}
    \label{fig:temporal_graph_multisimo}
    \vspace{-4mm}
\end{figure}

To investigate what determines task difficulty across segments, we computed correlations between turn-taking characteristics and human performance in \cref{tab:correlation}.
We considered four turn-taking-related features:(1) Turn-change probability, i.e., the proportion of adjacent utterance pairs in which the speaker changes within a one-minute window;
(2) Speaker dispersion, i.e., the entropy of utterance counts across speakers;
(3) Mean gap, i.e., the average temporal gap between consecutive utterances; and
(4) Silence ratio, i.e., the total duration of positive gaps (silences) normalized by the window duration.
Turn-change probability had a weak negative correlation with both addressee detection and next speaker prediction, indicating that conversations with more frequent speaker transitions tend to be more difficult.
Speaker dispersion also exhibited a weak negative correlation with performance, suggesting that conversations with evenly distributed participation are harder, whereas those dominated by a single speaker are relatively easier.
This tendency was more pronounced for next speaker prediction.
In contrast, mean gap and silence ratio showed no clear correlation with any task.
Turn-change prediction did not exhibit meaningful correlations with any of these features.
\textbf{Although these observations show that the difficulty of turn-taking in MPC is related to structural factors such as speaker distribution and the frequency of speaker transitions, these factors alone are not sufficient to fully explain performance variability.}
A variety of factors, such as discourse function, topic, personality, and social context, are considered to influence turn-taking~\cite{Sacks1974-ff,cox2025social,onishi2025modeling}, and further investigation is required to clarify how these factors relate to task difficulties.

\subsection{Pilot study on three-party conversation}
Our experiments used only the AMI corpus, and the findings may therefore be domain-dependent.
As a pilot study, we applied our evaluation to the Multisimo~\cite{Koutsombogera2018-ne}, a multimodal three-party conversation corpus.
We analyzed the first \mbox{$\sim$60} utterances of the S02 session for which addressee annotations were available.
\cref{fig:temporal_graph_multisimo} visualizes temporal changes for humans and Gemini 2.5 Pro performance using 15-second windows.
Although this analysis is small-scale and limited to one session, the trends of model and human performance appear similar to those observed on AMI.
Extending the evaluation to broader domains and languages remains future work, but this pilot result suggests that some of our findings may generalize beyond AMI.

\begin{table}[t]
\centering
\footnotesize
\setlength{\tabcolsep}{4pt}
\renewcommand{\arraystretch}{1.1}
\caption{Pearson Correlation r between turn-taking-related features (per one-minute window) and human performance.}
\vspace{-3mm}
\begin{tabular}{lrrr}
\toprule
\textbf{Characteristics} & \textbf{Addr.} & \textbf{Turn} & \textbf{Next} \\
\midrule
Turn-change probability & -0.27 & -0.08 & -0.29  \\
Speaker dispersion & -0.23 & 0.00 & -0.39 \\
Mean gap & -0.02 & 0.07 & 0.06 \\
Silence ratio & -0.14 & 0.04 & -0.14 \\
\bottomrule
\end{tabular}
\label{tab:correlation}
\vspace{-4mm}
\end{table}

\section{Limitation} \label{sec:limitation}
Our evaluation differs from natural conversational participation in several respects.
First, humans and MM-LLMs performed the tasks by watching fixed-angle recorded videos, which do not reflect the first-person perspective of a meeting participant.
Second, textual transcripts and explicit current speaker information were provided.
Such information is rarely available in natural interactions and may have led humans to rely more on linguistic cues than on visual and audio signals.
Third, some experimental parts began from the start of the meeting, while others began mid-session.
As a result, the amount of accessible conversational context varied across segments, potentially affecting difficulty.
Finally, although all subjects use English in professional settings, they were non-native speakers.
This gap between the experimental setting and real-world interaction may have led to an underestimation of human performance.

In addition, our experiments were conducted solely on the AMI corpus, which consists of role-based scenario meetings with four participants.
Different languages, domains, or interaction settings may yield different results.
Evaluating turn-taking abilities under more natural participation conditions and across diverse conversational settings remains for future research.

\section{Conclusion}
We conducted a unified evaluation of turn-taking in multimodal MPCs.
We compared supervised models, text-based LLMs, multimodal LLMs, and human participants on addressee detection, turn-change prediction, and next speaker prediction under online constraints.
Our results showed that multimodal LLMs underperformed humans in addressee detection and turn-change prediction, suggesting current limitations in effectively leveraging raw audio and visual signals without task- or domain-specific adaptation.
In contrast, text-based LLMs outperformed humans in next speaker prediction despite lacking access to multimodal information, suggesting strong contextual reasoning over conversational history.
These findings highlight both the progress and the limitations of current models in handling MPCs.
Future work should investigate how multimodal LLMs can more effectively integrate audio, video, and text information to better understand MPCs.
For practical deployment, it will also be necessary to develop systems that operate jointly with multi-talker automatic speech recognition and speaker diarization in realistic environments.
Finally, further investigation of human turn-taking behavior under comparable conditions will help inform the design and evaluation of MPC agents.
\newpage
\section{Generative AI Use Disclosure}
This manuscript was edited and polished with the assistance of generative AI.
Generative AI models were also used as comparison systems in the experimental evaluation.
All experimental design, implementation, and analysis were conducted by the authors who take full responsibility for the content.

\bibliographystyle{IEEEtran}
\bibliography{mybib,reference}

\end{document}